\documentclass[11pt,a4paper]{article}
\usepackage[hyperref]{conll2018}
\usepackage[caption=false]{subfig}
\usepackage{times,graphicx}
\usepackage{amsmath}

\usepackage{url}

\usepackage{booktabs}
\usepackage[dvipsnames]{colortbl}
\usepackage{amssymb}
\usepackage{graphicx}
\usepackage{tikz}
\usepackage{colortbl}

\usepackage[disable]{todonotes} 

\aclfinalcopy 

\newcommand\confname{CoNLL--SIGMORPHON}

\title{Copenhagen at \confname{} 2018: Multilingual Inflection in Context with Explicit Morphosyntactic Decoding}

\author{Yova Kementchedjhieva \\
  University of Copenhagen \\
  {\tt yova@di.ku.dk} \\\And
  Johannes Bjerva \\
  University of Copenhagen \\
  {\tt bjerva@di.ku.dk} \\ \And
  Isabelle Augenstein \\
  University of Copenhagen \\
  {\tt augenstein@di.ku.dk} \\}

\date{}

\begin{document}
\maketitle
\begin{abstract}
This paper documents the Team Copenhagen system which placed first in the CoNLL--SIGMORPHON 2018 shared task on universal morphological reinflection, Task 2 with an overall accuracy of 49.87.
Task 2 focuses on morphological inflection in context: generating an inflected word form, given the lemma of the word and the context it occurs in. Previous SIGMORPHON shared tasks have focused on context-agnostic inflection---the ``inflection in context'' task was introduced this year.
We approach this with an encoder-decoder architecture over character sequences with three core innovations, all contributing to an improvement in performance: (1) a wide context window; (2) a multi-task learning approach with the auxiliary task of MSD prediction; (3) training models in a multilingual fashion.
\end{abstract}

\section{Introduction}

This paper describes our approach and results for Task 2 of the CoNLL--SIGMORPHON 2018 shared task on universal morphological reinflection \cite{cotterell-conll-sigmorphon2018}. The task is to generate an inflected word form given its lemma and the context in which it occurs. 

Morphological (re)inflection from context is of particular relevance to the field of computational linguistics: it is compelling to estimate how well a machine-learned system can capture the morphosyntactic properties of a word given its context, and map those properties to the correct surface form for a given lemma. 

There are two tracks of Task 2 of CoNLL--SIGMORPHON 2018: in Track 1 the context is given in terms of word forms, lemmas and morphosyntactic descriptions (MSD); in Track 2 only word forms are available. See Table~\ref{tab:example} for an example. Task 2 is additionally split in three settings based on data size: high, medium and low, with high-resource datasets consisting of up to 70K instances per language, and low-resource datasets consisting of only about 1K instances. 

\begin{table*}
\resizebox{\linewidth}{!}{
\begin{tabular}{ c c c c c c c c c c}
 { \sc word forms} & We & were & \color{cyan} $\Box$ & to & feel & very & welcome & . \\ 
 \color{darkgray}{\sc lemmas} &\color{darkgray} we & \color{darkgray} be & make & \color{darkgray} to & \color{darkgray} feel & \color{darkgray} very & \color{darkgray} welcome & \color{darkgray} . \\
 \color{darkgray}\sc MSD tags &\color{darkgray} PRO;NOM;PL;1 & \color{darkgray} AUX;IND;PST;FIN & \color{magenta} $\Box$ & & \color{darkgray} PART & \color{darkgray} V;NFIN & \color{darkgray} ADV & \color{darkgray} ADJ & \color{darkgray} PUNCT  \\
\end{tabular}
\label{tab:example}
}
\caption{Example input sentence. Context MSD tags and lemmas, marked in gray, are only available in Track 1. The cyan square marks the main objective of predicting the word form \textit{made}. The magenta square marks the auxiliary objective of predicting the MSD tag \textit{V;PST;V.PTCP;PASS}.}
\end{table*}


The baseline provided by the shared task organisers is a \textit{seq2seq} model with attention (similar to the winning system for reinflection in CoNLL--SIGMORPHON 2016, \citet{kann2016med}), which receives information about context through an embedding of the two words immediately adjacent to the target form. We use this baseline implementation as a starting point and achieve the best overall accuracy of 49.87 on Task 2 by introducing three augmentations to the provided baseline system: (1) We use an LSTM to encode the entire available context; (2) We employ a multi-task learning approach with the auxiliary objective of MSD prediction; and (3) We train the auxiliary component in a multilingual fashion, over sets of two to three languages. 

In analysing the performance of our system, we found that encoding the full context improves performance considerably for all languages: 11.15 percentage points on average, although it also highly increases the variance in results. Multi-task learning, paired with multilingual training and subsequent monolingual finetuning, scored highest for five out of seven languages, improving accuracy by another 9.86\% on average. 

\section{System Description}

Our system is a modification of the provided CoNLL--SIGMORPHON 2018 baseline system, so we begin this section with a reiteration of the baseline system architecture, followed by a description of the three augmentations we introduce.

\subsection{Baseline}
The CoNLL--SIGMORPHON 2018 baseline\footnote{Code available at:\\ \url{https://github.com/sigmorphon/conll2018}} is described as follows: 
\begin{quote}
The system is an encoder-decoder on character sequences. It takes a lemma as input and generates a word form. The process is conditioned on the context of the lemma [\ldots] The baseline treats the lemma, word form and MSD of the previous and following word as context in track 1. In track 2, the baseline only considers the word forms of the previous and next word. [\ldots] The baseline system concatenates embeddings for context word forms, lemmas and MSDs into a context vector. The baseline then computes character embeddings for each character in the input lemma. Each of these is concatenated with a copy of the context vector. The resulting sequence of vectors is encoded using an LSTM encoder. Subsequently, an LSTM decoder generates the characters in the output word form using encoder states and an attention mechanism.
\end{quote}

\begin{figure}[tb]
\resizebox{\linewidth}{!}{
\includegraphics[]{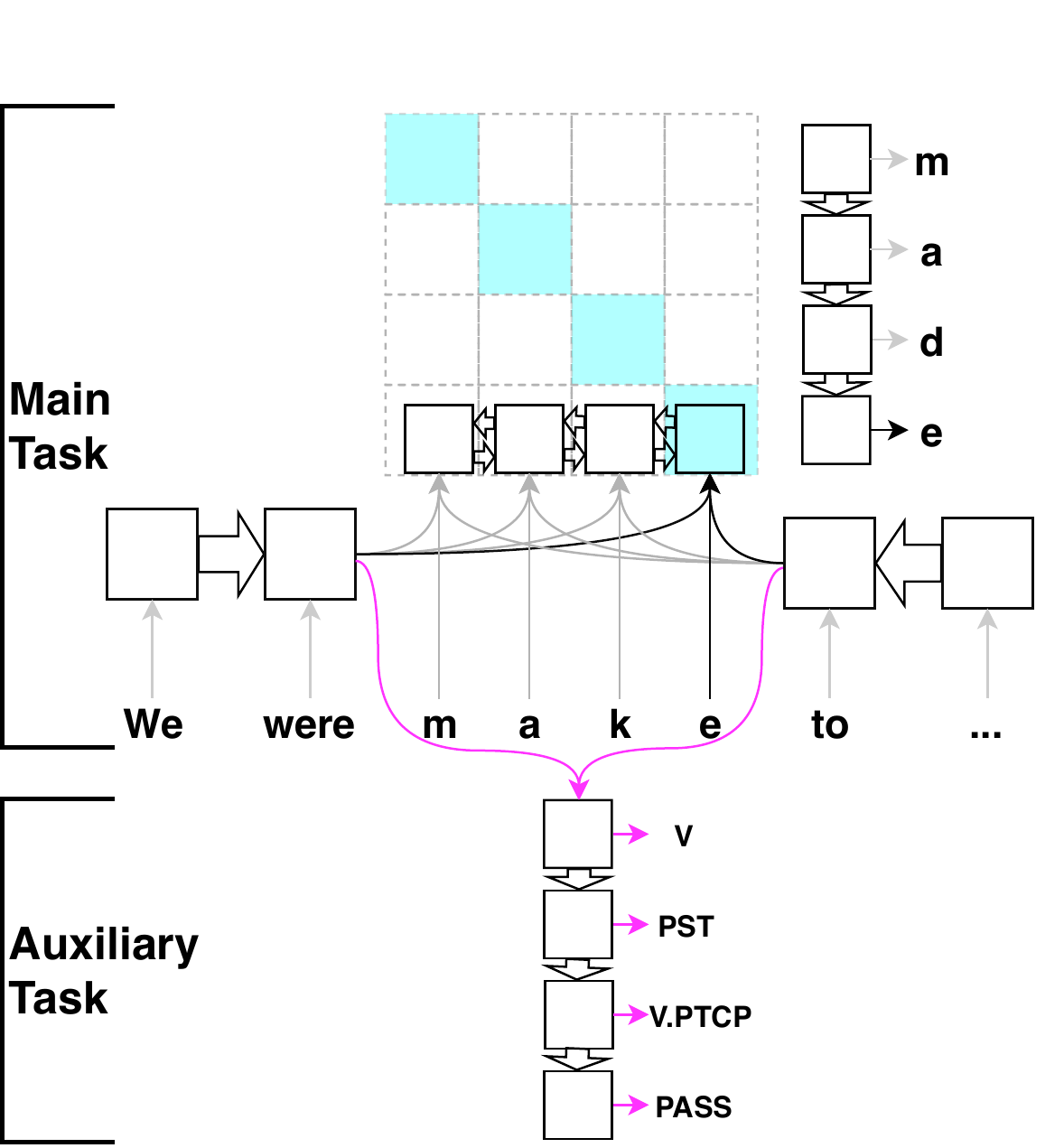}
} \caption{Schematic representation of our approach. The focus here is on the prediction of the final character, \textit{e}, of the word form \textit{made}. The attention matrix indicates that this character should be based on the final state of the encoder, which contains information about the final character of the input form, and the past and future context. The input and output of the auxiliary decoder are marked in magenta.}
\label{fig:architecture}
\end{figure}

To that we add a few details regarding model size and training schedule:
\begin{itemize}
\itemsep0em 
\item the number of LSTM layers is one;
\item embedding size, LSTM layer size and attention layer size is 100;
\item models are trained for 20 epochs;
\item on every epoch, training data is subsampled at a rate of 0.3;
\item LSTM dropout is applied at a rate 0.3;
\item context word forms are randomly dropped at a rate of 0.1;
\item the Adam optimiser is used, with a default learning rate of 0.001; and
\item trained models are evaluated on the development data (the data for the shared task comes already split in train and dev sets).
\end{itemize}


\subsection{Our system}

Here we compare and contrast our system\footnote{Code available at: \url{https://github.com/YovaKem/inflection_in_context}} to the baseline system. A diagram of our system is shown in Figure \ref{fig:architecture}.

\subsubsection{Entire Context Encoded with LSTMs}\label{context_encoding}

The idea behind this modification is to provide the encoder with access to all morpho-syntactic cues present in the sentence.
In contrast to the baseline, which only encodes the immediately adjacent context of a target word, we encode the \textit{entire context}. All context word forms, lemmas, and MSD tags (in Track 1) are embedded in their respective high-dimensional spaces as before, and their embeddings are concatenated. However, we now reduce the entire past context to a fixed-size vector by encoding it with a forward LSTM, and we similarly represent the future context by encoding it with a backwards LSTM. 

\subsubsection{Auxiliary Task: MSD of the Target Form}

We introduce an auxiliary objective that is meant to \textit{increase the morpho-syntactic awareness of the encoder} and to regularise the learning process---the task is to predict the MSD tag of the target form. 
MSD tag predictions are conditioned on the context encoding, as described in \ref{context_encoding}. Tags are generated with an LSTM one component at a time, e.g.\ the tag \textit{PRO;NOM;SG;1} is predicted as a sequence of four components, $\langle$PRO, NOM, SG, 1$\rangle$. 

For every training instance, we backpropagate the sum of the main loss and the auxiliary loss without any weighting.

As MSD tags are only available in Track 1, this augmentation only applies to this track. 

\begin{table}[tb]
\centering
\begin{tabular}{llrr|rr}
\toprule
\multicolumn{2}{c}{}& \multicolumn{2}{c|}{Track 1} & \multicolumn{2}{c}{Track 2}  \\ 
 \multicolumn{2}{c}{} & \multicolumn{1}{l}{base} & \multicolumn{1}{l|}{our} & \multicolumn{1}{l}{base} & \multicolumn{1}{l}{our} \\ \midrule
 & \sc de & 64.51 & \textbf{72.40} & \textbf{65.72} & 64.81 \\ 
 & \sc en & 72.91 & \textbf{77.84} & 70.39 & \textbf{71.90} \\ 
 &\sc es & 53.44 & \textbf{56.24} & \textbf{51.05} & 48.95 \\ 
high & \sc fi & 49.05 & \textbf{55.27} & \textbf{34.82} & 32.40 \\ 
 & \sc fr & 63.54 & \textbf{70.67} & 58.45 & \textbf{61.51} \\ 
 & \sc ru & 71.18 & \textbf{77.91} & 46.89 & \textbf{49.00} \\ 
 & \sc sv & 62.23 & \textbf{69.26} & 54.04 & \textbf{55.96} \\ \midrule
 & \sc de & 54.40 & \textbf{62.18} & 56.93 & \textbf{57.33} \\ 
 & \sc en & 60.02 & \textbf{66.67} & 57.60 & \textbf{66.67} \\ 
 & \sc es & 23.14 & \textbf{51.33} & 41.23 & \textbf{42.50} \\ 
med. & \sc fi & 28.21 & \textbf{35.71} & 19.19 & \textbf{22.24} \\ 
 & \sc fr & 45.01 & \textbf{60.29} & 21.38 & \textbf{45.62} \\ 
 & \sc ru & 50.30 & \textbf{63.05} & 30.52 & \textbf{35.94} \\ 
 & \sc sv & 47.55 & \textbf{57.66} & 43.09 & \textbf{45.96} \\ \midrule
& \sc de & 0.20 & \textbf{4.85} & 0.10 & \textbf{18.91} \\ 
 & \sc en & 1.81 & \textbf{33.84} & 2.22 & \textbf{59.42} \\ 
 & \sc es & 8.98 & \textbf{31.42} & 8.98 & \textbf{31.84} \\ 
low & \sc fi & 0.76 & \textbf{12.83} & 0.38 & \textbf{12.33} \\ 
 & \sc fr & 0.00 & \textbf{34.42} & 0.00 & \textbf{29.53} \\ 
 & \sc ru & 0.00 & \textbf{25.90} & 2.71 & \textbf{22.69} \\ 
 & \sc sv & 1.17 & \textbf{27.55} & 0.96 & \textbf{30.96} \\ \bottomrule
\end{tabular}
\caption{Official shared task test set results.}
\label{test_results}
\end{table}

\subsubsection{Multilinguality}

The parameters of the entire MSD (auxiliary-task) decoder are shared across languages. 

Since a grouping of the languages based on language family would have left several languages in single-member groups (e.g.\ Russian is the sole representative of the Slavic family), we experiment with random \textit{groupings of two to three languages}. Multilingual training is performed by randomly alternating between languages for every new minibatch. We do not pass any information to the auxiliary decoder as to the source language of the signal it is receiving, as we assume abstract morpho-syntactic features are shared across languages.

\paragraph{Finetuning} After 20 epochs of multilingual training, we perform 5 epochs of monolingual finetuning for each language. For this phase, we reduce the learning rate to a tenth of the original learning rate, i.e.\ 0.0001, to ensure that the models are indeed being finetuned rather than retrained.

\subsubsection{Model Size and Training Schedule}
We keep all hyperparameters the same as in the baseline. 
Training data is split 90:10 for training and validation. We train our models for 50 epochs, adding early stopping with a tolerance of five epochs of no improvement in the validation loss. We do not subsample from the training data. 

\subsubsection{Ensemble Prediction}

We train models for 50 different random combinations of two to three languages in Track 1, and 50~monolingual models for each language in Track~2. Instead of picking the single model that performs best on the development set and thus risking to select a model that highly overfits that data, we use an ensemble of the five best models, and make the final prediction for a given target form with a majority vote over the five predictions. 

\section{Results and Discussion}

Test results are listed in Table \ref{test_results}. 
Our system outperforms the baseline for all settings and languages in Track 1 and for almost all in Track 2---only in the high resource setting is our system not definitively superior to the baseline. 

Interestingly, our results in the low resource setting are often higher for Track 2 than for Track 1, even though contextual information is less explicit in the Track 2 data and the multilingual multi-tasking approach does not apply to this track. We interpret this finding as an indicator that a simpler model with fewer parameters works better in a setting of limited training data. Nevertheless, we focus on the low resource setting in the analysis below due to time limitations. As our Track 1 results are still substantially higher than the baseline results, we consider this analysis valid and insightful. 

\subsection{Ablation Study} 

\begin{figure*}[tb]
\resizebox{\textwidth}{!}{
\includegraphics[]{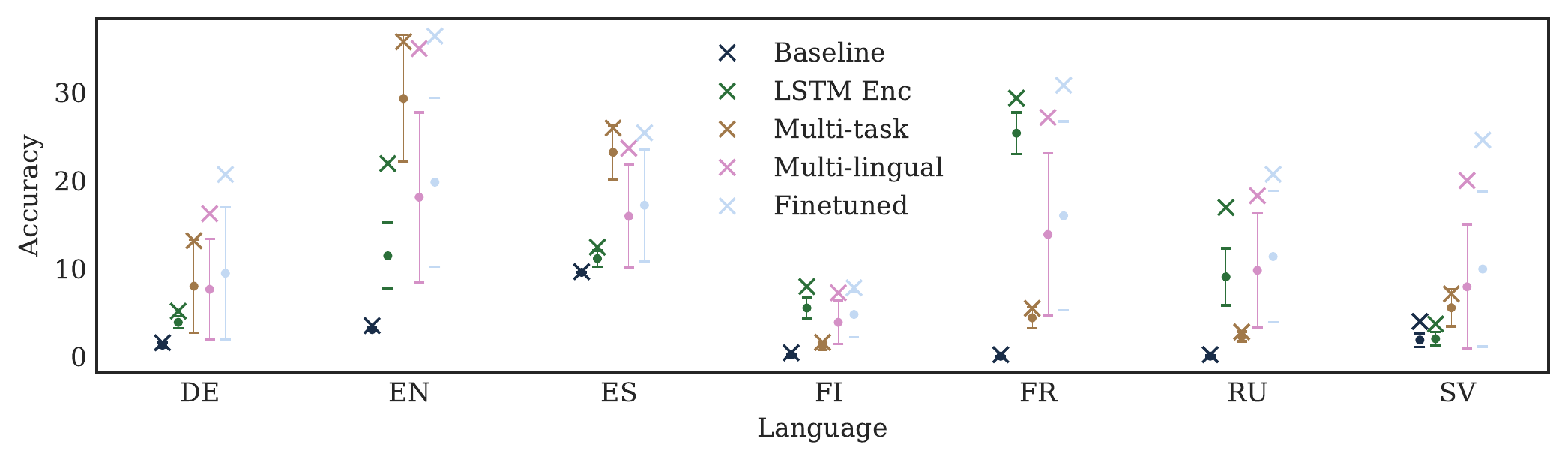}
} \caption{Mean ($\bullet$) and standard deviation (error bars)  over 100 models trained for each language and architecture, and average ($\times$) over the 5 best models. \textit{LSTM Enc} refers to a model that encodes the full context with an LSTM; \textit{Multi-task} builds on \textit{LSTM Enc} with an auxiliary objective of MSD prediction; \textit{Multilingual} refers to a model with an auxiliary component trained in a multilingual fashion; \textit{Finetuned} refers to a multilingual model topped with monolingual finetuning. }
\label{fig:results}
\end{figure*}

We analyse the incremental effect of the different features in our system, focusing on the low-resource setting in Track 1 and using development data. 

\paragraph{Entire Context Encoded with LSTMs}
Encoding the entire context with an LSTM highly increases the variance of the observed results. So we trained fifty models for each language and each architecture. Figure~\ref{fig:results} visualises the means and standard deviations over the trained models. 
In addition, we visualise the average accuracy for the five best models for each language and architecture, as these are the models we use in the final ensemble prediction. Below we refer to these numbers only.

The results indicate that encoding the full context with an LSTM highly enhances the performance of the model, by 11.15\% on average. This observation explains the high results we obtain also for Track 2.  

\paragraph{Auxiliary Task: MSD of the Target Form}
Adding the auxiliary objective of MSD prediction has a variable effect: for four languages ({\sc de}, {\sc en}, { \sc es}, and {\sc sv}) the effect is positive, while for the rest it is negative. We consider this to be an issue of insufficient data for the training of the auxiliary component in the low resource setting we are working with.

\paragraph{Multilinguality}
We indeed see results improving drastically with the introduction of multilingual training, with multilingual results being 7.96\% higher than monolingual ones on average. 

We studied the five best models for each language as emerging from the multilingual training (listed in Table~\ref{tab:5best}) and found no strong linguistic patterns.
The {\sc en}--{\sc sv} pairing seems to yield good models for these languages, which could be explained in terms of their common language family and similar morphology. The other natural pairings, however, {\sc fr}--{\sc es}, and {\sc de}--{\sc sv}, are not so frequent among the best models for these pairs of languages. 

\begin{table}[tb]
\resizebox{\linewidth}{!}{
\begin{tabular}{|l|lllll|}
\toprule
\sc de&\sc fi & \sc sv& \sc fi, \sc sv &  \sc ru, \sc fr & \sc fr, \sc fi \\ 
\sc en&\sc ru, \sc sv & \sc ru, \sc fi & \sc ru,\sc fr & \sc sv, \sc es & \sc sv, \sc fr \\ 
\sc es& \sc de  & \sc fi &\sc sv, \sc de & \sc sv,\sc en & \sc sv,\sc fr\\ 
\sc fi&\sc de  & \sc es & \sc fr, \sc es & \sc en,\sc ru & \sc ru,\sc sv \\ 
\sc fr&\sc sv,\sc en & \sc en,\sc es& \sc de,\sc fi & \sc sv,\sc en & \sc en,\sc sv \\ 
\sc ru& \sc sv & \sc de,\sc fr & \sc en,\sc sv & \sc sv,\sc fr & \sc en,\sc fi \\ 
\sc sv&\sc en,\sc de &\sc fi,\sc en &\sc fr,\sc ru & \sc es,\sc en & \sc ru, \sc en \\ \bottomrule
\end{tabular}
}
\caption{\label{tab:5best}Five best multilingual models for each language.}
\end{table}

Finally, monolingual finetuning improves accuracy across the board, as one would expect, by 2.72\% on average. 

\paragraph{Overall}
The final observation to be made based on this breakdown of results is that the multi-tasking approach paired with multilingual training and subsequent monolingual finetuning outperforms the other architectures for five out of seven languages: {\sc de}, {\sc en}, {\sc fr}, {\sc ru} and {\sc sv}. For the other two languages in the dataset, {\sc es} and {\sc fi}, the difference between this approach and the approach that emerged as best for them is less than~1\%. The overall improvement of the multilingual multi-tasking approach over the baseline is~18.30\%.

\subsection{Error analysis}
Here we study the errors produced by our system on the English test set to better understand the remaining shortcomings of the approach. A small portion of the wrong predictions point to an incorrect interpretation of the morpho-syntactic conditioning of the context, e.g.\ the system predicted \textit{plan} instead of \textit{plans} in the context \textit{Our \_ include raising private capital}. The majority of wrong predictions, however, are nonsensical, like \textit{bomb} for \textit{job}, \textit{fify} for \textit{fixing}, and \textit{gnderrate} for \textit{understand}. This observation suggests that generally the system did not learn to copy the characters of lemma into inflected form, which is all it needs to do in a large number of cases. This issue could be alleviated with simple data augmentation techniques that encourage autoencoding \citep[see, e.g.,][]{bergmanis2017training}.

\subsection{MSD prediction}

\begin{figure}
\resizebox{\linewidth}{!}{
\includegraphics[]{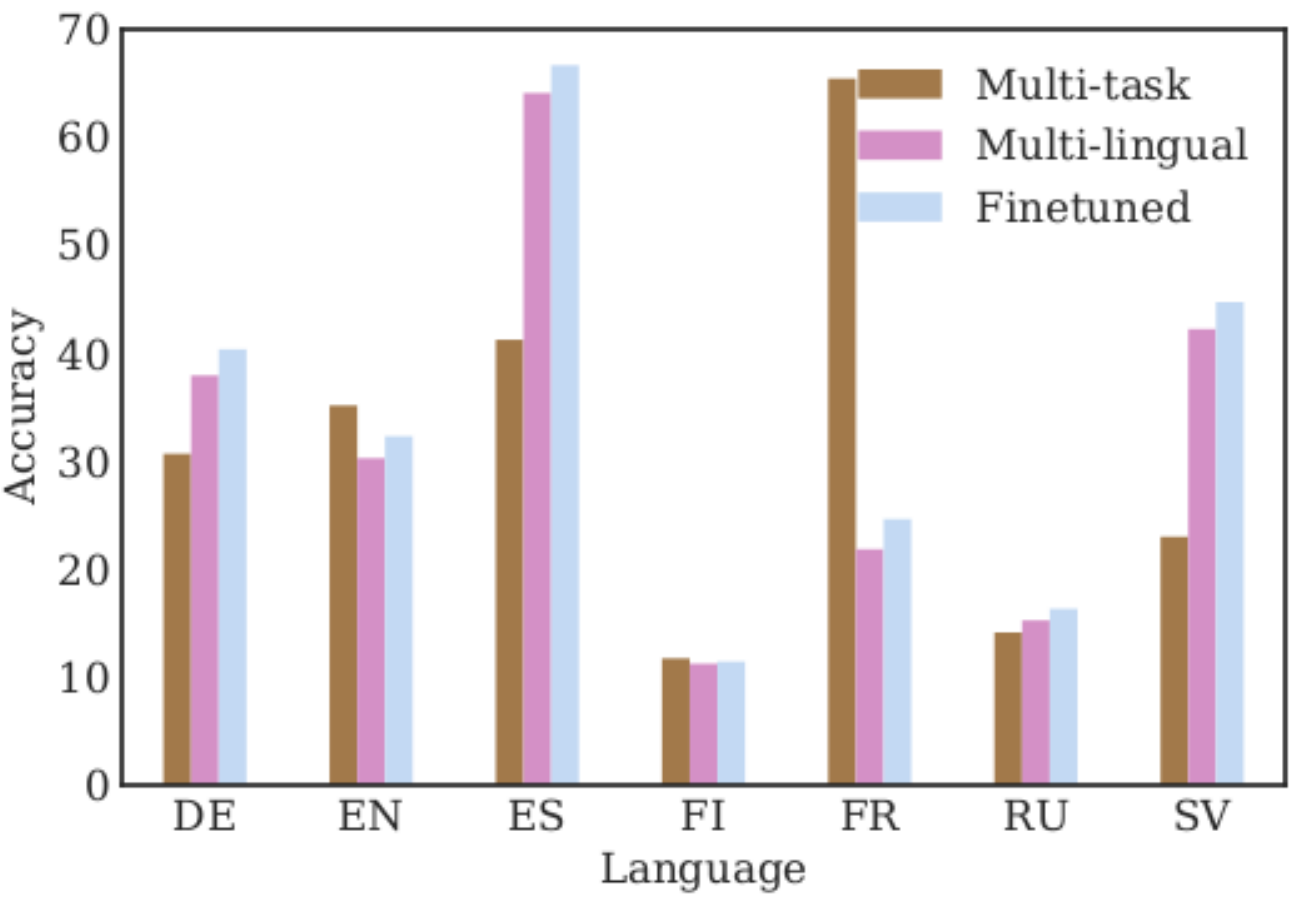}
} \caption{Accuracy on the auxiliary task of MSD prediction with different models. See the caption of Figure~\ref{fig:results} for more details.}
\label{fig:results_msd}
\end{figure}

Figure~\ref{fig:results_msd} summarises the average MSD-prediction accuracy for the multi-tasking experiments discussed above.\footnote{As MSD tags are not available for target forms in the development data, the accuracy of MSD prediction is measured over all other nouns, adjectives and verbs in the dataset.} Accuracy here is generally higher than on the main task, with the multilingual finetuned setup for Spanish and the monolingual setup for French scoring best: 66.59\% and 65.35\%, respectively. This observation illustrates the added difficulty of generating the correct surface form even when the morphosyntactic description has been identified correctly.  


We observe some correlation between these numbers and accuracy on the main task: for {\sc de, en, ru} and {\sc sv}, the brown, pink and blue bars here pattern in the same way as the corresponding $\times$'s in Figure~\ref{fig:results}. One notable exception to this pattern is {\sc fr} where inflection gains a lot from multilingual training, while MSD prediction suffers greatly. Notice that the magnitude of change is not always the same, however, even when the general direction matches: for {\sc ru}, for example, multilingual training benefits inflection much more than in benefits MSD prediction, even though the MSD decoder is the only component that is actually shared between languages. This observation illustrates the two-fold effect of multi-task training: an auxiliary task can either inform the main task through the parameters the two tasks share, or it can help the main task learning through its regularising effect. 

\section{Related Work}
Our system is inspired by previous work on multi-task learning and multi-lingual learning, mainly building on two intuitions:
(1)~jointly learning related tasks tends to be beneficial \cite{mtl,sogaard2016deep,plank:2016,bjerva:2016:semantic,bjerva:2017:mtl}; and
(2)~jointly learning related languages in an MTL-inspired framework tends to be beneficial \citep{bjerva:phd,google:zeroshot,delhoneux:2018}.
In the context of computational morphology, multi-lingual approaches have previously been employed for morphological reinflection \citep{bergmanis2017training} and for paradigm completion \citep{kann2017one}.
In both of these cases, however, the available datasets covered more languages, 40 and~21, respectively, which allowed for linguistically-motivated language groupings and for parameter sharing directly on the level of characters. 
\Citet{delhoneux:2018} explore parameter sharing between related languages for dependency parsing, and find that sharing is more beneficial in the case of closely related languages.

\section{Conclusions}
In this paper we described our system for the CoNLL--SIGMORPHON 2018 shared task on Universal Morphological Reinflection, Task 2, which achieved the best performance out of all systems submitted, an overall accuracy of 49.87.
We showed in an ablation study that this is due to three core innovations, which extend a character-based encoder-decoder model: (1)~a wide context window, encoding the entire available context; (2)~multi-task learning with the auxiliary task of MSD prediction, which acts as a regulariser; (3)~a multilingual approach, exploiting information across languages.
In future work we aim to gain better understanding of the increase in variance of the results introduced by each of our modifications and the reasons for the varying effect of multi-task learning for different languages.

\section*{Acknowledgements}
We gratefully acknowledge the support of the NVIDIA Corporation with the donation of the Titan Xp GPU used for this research.

\bibliography{conll2018}
\bibliographystyle{acl_natbib_nourl}

\end{document}